\documentclass[10pt,twocolumn,letterpaper]{article}

\usepackage{cvpr}              

\definecolor{cvprblue}{rgb}{0.21,0.49,0.74}
\usepackage[pagebackref,breaklinks,colorlinks,allcolors=cvprblue]{hyperref}

\def\paperID{*****} 
\def\confName{CVPR}
\def\confYear{2025}

\title{Adding Another Dimension to Image-based Animal Detection}

\author{
Vandita Shukla\\
Fondazione Bruno Kessler\\
Trento, Italy\\
{\tt\small vshukla@fbk.eu}
\and
Fabio Remondino\\
Fondazione Bruno Kessler\\
Trento, Italy\\
{\tt\small remondino@fbk.eu}
\and
Benjamin Risse\\
University of Muenster\\
Muenster, Germany\\
{\tt\small b.risse@uni-muenster.de}
}

\begin{document}
\maketitle
\begin{abstract}
Monocular imaging of animals inherently reduces 3D structures to 2D projections.
Detection algorithms lead to 2D bounding boxes that lack information about animal’s orientation relative to the camera.
To build 3D detection methods for RGB animal images, there is a lack of labeled datasets; such labeling processes require 3D input streams along with RGB data.
We present a pipeline that utilizes Skinned Multi-Animal Linear models to estimate 3D bounding boxes and to project them as robust labels into 2D image space using a dedicated camera pose refinement algorithm.
To assess which sides of the animal are captured, cuboid face visibility metrics are computed.
These 3D bounding boxes and metrics form a crucial step toward developing and benchmarking future monocular 3D animal detection algorithms.
We evaluate our method on the Animal3D dataset, demonstrating accurate performance across species and settings.
\end{abstract}    
\section{Introduction and related work}
\label{sec:intro_workshop}

Animal biometrics is a research field focused on quantifying phenotypical characteristics and their changes in individual animals \cite{kuhl_animal_2013, farrag_biometrics_2022}.
Visual animal biometrics (VAB) has emerged as a key technique for identifying and tracking animals based on unique physical features, with applications in conservation, livestock management, and anti-poaching \cite{kumar_cattle_2020, kumar_visual_2017}.

Recent advances in computer vision and machine learning have enabled non-invasive VAB, eliminating the need for tagging or microchipping \cite{langley_assessing_2022,cheeseman_advanced_2022}.
However, camera-based imaging, such as camera trap recordings, inherently reduces 3D objects to 2D projections. 
Standard object detection algorithms extract 2D bounding boxes, which help localize the animals in image-space but lack geometric context \cite{petso_review_2022}.
Image-based animal detection, without characterizing the visible orientation of animal profiles in the image, limits both the observation and utilization of VAB features, as
(1) appearances may vary significantly, for example, between left and right profile views \cite{RAVOOR2020100289}; and
(2) VAB features are related to specific profiles, such as facial \cite{clapham_multispecies_2022, gallo_individual_2022} or side profile \cite{dheer_using_2022}.
This motivates a reevaluation of current 2D detection-based monitoring approaches, as for a more complete phenotypical profiling, it is crucial to associate visual features with the corresponding side of the animal. 

\begin{figure}[!tb]
  \centering
   \includegraphics[width=0.8\linewidth]{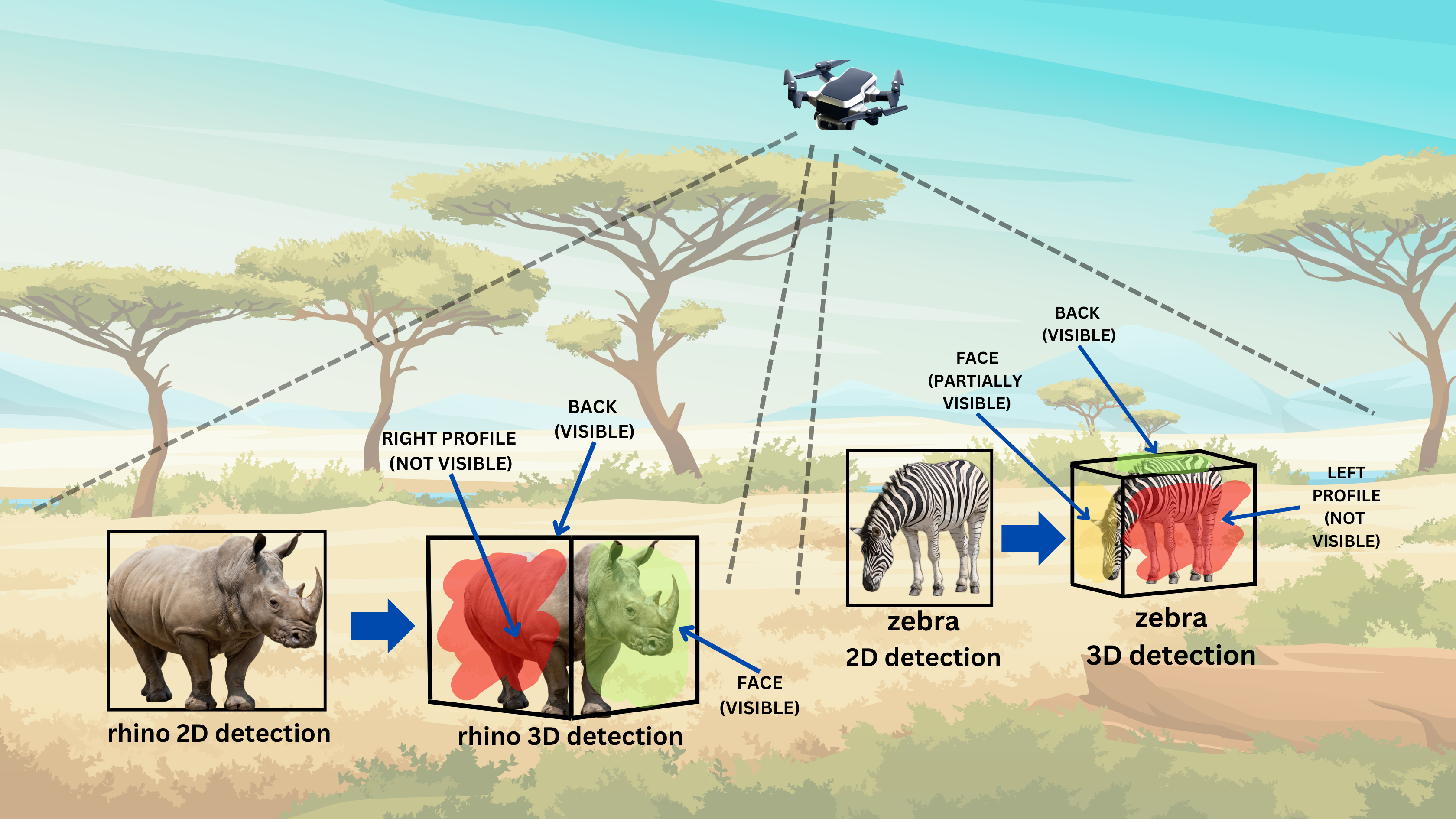}
   \caption{Illustrative description of lifting 2D detection to oriented 3D-bounding boxes.}
   \label{fig:profile_decon}
\end{figure}

One solution is monocular 3D object detection, i.e., lifting 2D bounding boxes to 3D cuboids in image space (Figure \ref{fig:profile_decon}). 
This process performs not only 3D object detection but also monocular 6-degree-of-freedom (DoF) pose estimation for cuboid orientation and localization. 
These are challenging tasks in 3D scene understanding, with active research in autonomous driving, robotic manipulation, and AR \cite{ contreras_survey_2024, fan_deep_2022}.
Analogous to autonomous driving, wildlife monitoring using Unmanned Aerial Vehicles (UAV) is steadily evolving toward autonomous functionality \cite{kline_wildwing_2025, meier_wildbridge_2024}. 
In this context, image-based 3D animal detection could enable UAVs to autonomously reposition in real-time to capture optimal animal profile viewpoints, while also supporting downstream data analysis.

\begin{figure}[!tb]
  \centering
   \includegraphics[width=0.9\linewidth]{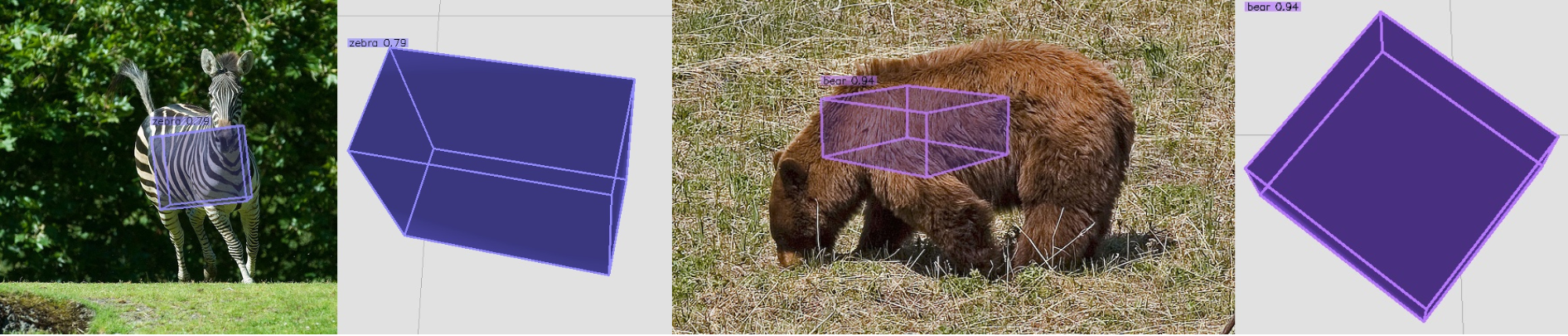}
   \caption{Zero shot predictions from Ovmono3D \cite{yao2024openvocabularymonocular3d} on Imagenet samples.}
   \label{fig:zeroshot}
\end{figure}

Applying 3D detection methods to animals, however, adds further complexity due to non-rigid shapes, diverse poses and appearances, and cluttered environments. 
3D annotation remains a major bottleneck; manually labeling 3D boxes in images is nearly impossible without auxiliary 3D data \cite{mouawad_view--label_2025, meng_towards_2022}. 
Current tools often rely on RGB-D sequences or LiDAR-enhanced point clouds \cite{stumpf_salt_2021-1,huang_training_2024}. 
Even newer purely RGB-based approaches tend to support only rigid objects of known shapes or almost always require manual perspective snapping, known camera intrinsics, or video sequences, impractical for in-the-wild animal images with arbitrary viewpoints \cite{simonelli_disentangling_2022, qu_monodcn_2022}. 
Introducing new object classes also suffers from a cold-start problem: without high-quality 3D bounding box annotations, it is difficult to train or validate models to assist labeling for monocular 3D animal detection (Figure \ref{fig:zeroshot}). 
Thus, leveraging existing curated RGB-only datasets to produce viewpoint consistent oriented 3D boxes remains highly challenging.

A promising avenue is to utilize Skinned Multi-Animal Linear (SMAL)\cite{zuffi_3d_2017} models as auxiliary 3D evidence, especially with the recent release of datasets like Animal3D \cite{xu_animal3d_2023}. 
However, directly projecting mesh-derived 3D boxes onto images often results in misaligned or incorrect orientations due to: 
(1) small manual annotation errors propagating into large pose estimation errors, 
(2) insufficient global constraints from 3D-to-2D correspondences, often leading to degenerate or depth-flipped boxes, 
(3) unreliable orientation estimates from ubiquitous methods like Principal Component Analysis (PCA) in the presence of complex poses and occlusion.


We propose a pipeline for generating high-quality 3D bounding box annotations while overcoming the above limitations and leveraging SMAL to override the need for multi-sensor data. 
The generated labels aim to support the development of monocular 3D detection pipelines and deep learning-assisted labeling workflows. 
To provide view-specific information for VAB applications, we augment the labels by introducing an animal profile visibility metric.
We evaluate our 3D bounding box label-generation pipeline on the Animal3D dataset and UAV-captured data, demonstrating robust label generation across diverse viewpoints. 

\section{Methodology}
\label{sec:prop_method}
\begin{figure}[!htb]
  \centering
   \includegraphics[width=0.85\linewidth]{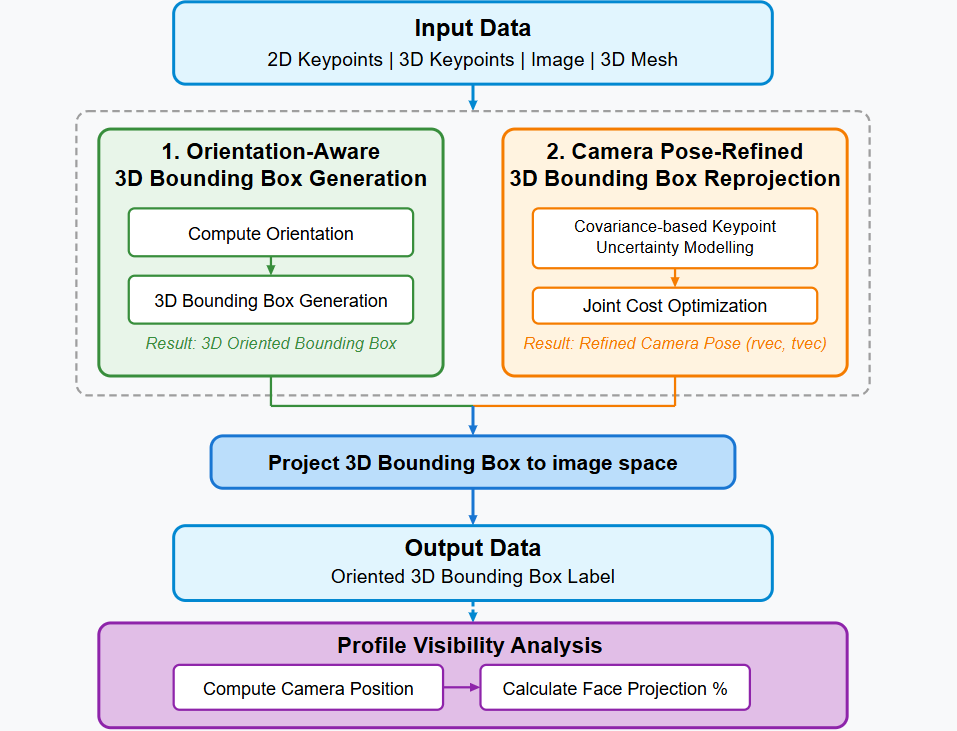}
   \caption{Pipeline overview. 
   Regarding inputs, it is important to clarify that the SMAL model fitting inherently requires 2D keypoint annotations as part of its standard pipeline. 
   These keypoints establish critical correspondence between the input image and 3D model, enabling the optimization process to infer 3D keypoints during fitting. 
   The 3D keypoints emerge naturally as a byproduct of the SMAL fitting process, rather than requiring separate generation as additional system inputs.}
   \label{fig:pipeline}
\end{figure}

We present an end-to-end pipeline that leverages SMAL models to generate 3D bounding box labels on images.
An overview of the methodology is shown in Figure \ref{fig:pipeline}.
Given an input image and corresponding SMAL model, the algorithm extracts an orientation-aware 3D bounding box and a refined camera pose. 
These are combined to form a 3D box representation in the 2D image along with the profile visibility metric. 
Due to the reliance on SMAL models as auxiliary 3D reference data for corresponding images, an inherent limitation is that the 3D bounding box labels can only be generated for the animal species that are representable within the SMAL feature space.

\subsection{Orientation-aware 3D bounding box generation}
\subsubsection{Orientation}
To generate correctly oriented 3D bounding boxes from SMAL, we use a targeted method for defining consistent bounding box axes across meshes. This was motivated by the failure of PCA to generalize across species and poses, due to:
(i) skew from non-representative features like tails, as PCA does not prioritize semantically meaningful regions;
(ii) axis ambiguity from symmetric structures, causing eigenvector flips; and
(iii) misalignment between mesh vertex distributions and the natural orientation of animals in complex poses.

To address this, we use 3D landmark keypoints and their image-space visibility to create a fallback mechanism for selecting stable axes.
The x-axis is defined along the anterior-posterior direction, the y-axis along the left-right direction, and the z-axis is derived using Fleming’s left-hand rule.

\subsubsection{3D Bounding Box Generation}
The bounding box dimensions and position are calculated using a standard coordinate system. 
Mesh vertices are converted to local coordinate system of the anatomically aligned rotation matrix $R_{\text{box}}^T$, using  
$
V_{\text{local}} = R_{\text{box}}^T (V_{\text{world}} - c)
$
where $
c = \frac{1}{N}\sum_{i=1}^{N} V_i
$ is the centroid.
The minimum and maximum coordinates of the box are calculated from the transformed mesh vertices using \[
(V_{\text{local}}^{\min},\,V_{\text{local}}^{\max}) = \Bigl(\min(V_{\text{local}})-\epsilon,\; \max(V_{\text{local}})+\epsilon\Bigr)
\] with an optional small margin $\epsilon$. 
We maintain this small margin to prevent false negatives caused by floating-point errors during mesh enclosure calculations for evaluation. 
We use $\epsilon = 10^{-5}$, which is negligible relative to typical mesh coordinates but ensures that vertices lying exactly on box boundaries are correctly classified as enclosed.
The 8 corners of the bounding box are generated in the local coordinate system
as $\{x_{\min},x_{\max}\} \times \{y_{\min},y_{\max}\} \times \{z_{\min},z_{\max}\}
$ where $(x_{\text{min}}, y_{\text{min}}, z_{\text{min}})$ and $(x_{\text{max}}, y_{\text{max}}, z_{\text{max}})$ are the components of $V_{\text{local}}^{\text{min}}$ and $V_{\text{local}}^{\text{max}}$, respectively. 
These are then transformed into the world coordinates using
$\text{corners}_{\text{world}} = R_{\text{box}} \cdot \text{corners}_{\text{local}} + c$

\subsection{Camera pose refinement for 3D bounding box reprojection} 
We used Efficient Perspective-n-Point (EPnP) algorithm \cite{lepetit_epnp_2009} to establish an initial camera pose by utilizing the 2D-3D correspondences between visible keypoints.  
RANSAC \cite{fischler_random_1981} iteratively samples minimal corresponding pairs to find a model with maximum inliers.  
This yields an initial rotation and translation vector:
\begin{equation*}
\min_{R, t} \sum_{i} \rho \left( \| P(R \cdot X_i + t) - x_i \| \right)
\end{equation*}
where $P$ is the projection function, $X_i$ are 3D keypoints, $x_i$ are 2D keypoints, and $\rho$ is the robust cost function RANSAC. 
Due to the inherent ill-posed nature of the problem ambiguous solutions through small keypoint errors cause "flips" where the depth appears mirrored or produce degenerate solutions where the animal ends up behind the camera. 
We thus introduce two refinement steps. 

\subsubsection{Covariance-based keypoint uncertainty modelling}
To improve our camera pose, we added keypoint uncertainty using covariance matrices to attach a metric of importance to each landmark. 
The covariance matrix that models the keypoint uncertainty for each keypoint is represented as:
\begin{equation*}\Sigma_i = \begin{bmatrix} 
\sigma_i^2 & 0 \\
0 & \sigma_i^2
\end{bmatrix}\end{equation*}
Here $\sigma_i^2$ is the variance determined by (1) $\sigma_{\text{base}}^2$ which is the visibility or occlusion of keypoints - visibility contributing to a lower variance and vice versa , (2) $\text{edge\_factor}$ which is the distance to image boundaries where lower distance represents both the possibility of distortion and the acknowledgement of partial visibility, and (3) $\text{conf\_factor}$, which is the keypoint confidence scores (if available).
This uncertainty model allows the subsequent optimization to naturally prioritize reliable keypoints where 
\begin{equation*}\sigma_i^2 = \sigma_{\text{base}}^2 \cdot \text{edge\_factor} \cdot \text{conf\_factor}
\end{equation*}

\begin{figure*}
    \centering
    \includegraphics[width=\textwidth]{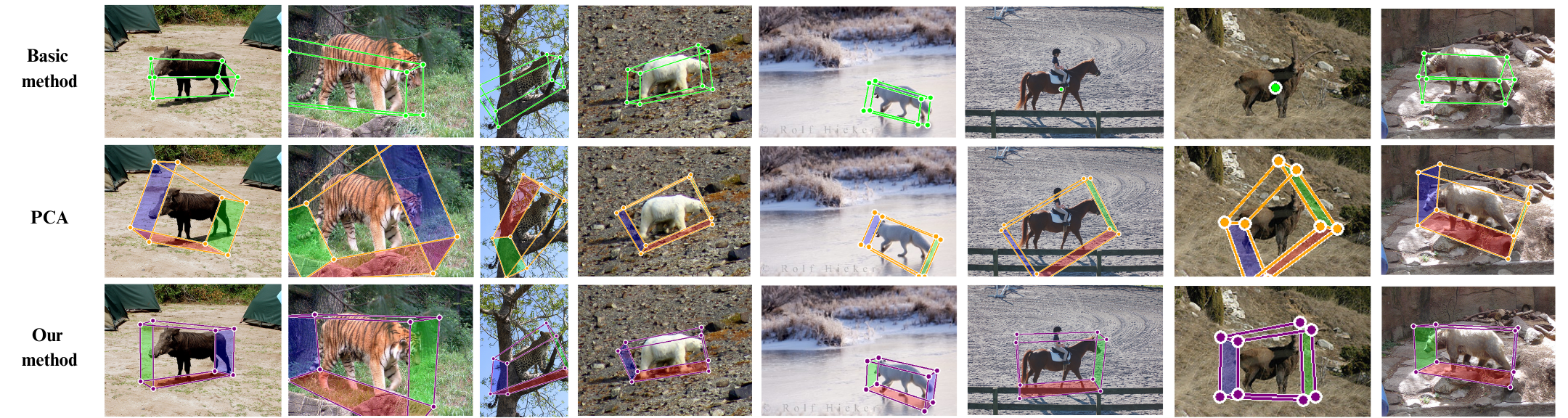}
    \caption{Reprojected bounding box  derived with Basic Method (top row), PCA (middle row) and Our Method (bottom row). 
    The proposed method shows improvement in both reprojection performance (Basic method vs Our Method) and in inferring meaningful orientation (PCA vs Our Method). 
    Note that green and blue faces represent the front and back profiles, respectively, which are randomly flipped using PCA.}
    \label{fig:orientation_ablation}
\end{figure*}

\subsubsection{Joint cost optimization}
While the keypoints help maintain the local anatomical precision, we use a segmentation mask to add global constraint ensuring overall spatial correctness of the animal. 
This segmentation mask can be derived from the superimposition of SMAL meshes on the respective image. 
Our residual function jointly optimizes camera pose by combining keypoint reprojection errors with mask alignment constraints, improving stability in case of high keypoint uncertainty.
The combined objective function is as follows:
 \begin{equation*}E_{\text{total}} = \lambda \sum_{i} d_{\text{Mahalanobis}}(x_i, \hat{x}_i, \Sigma_i) + (1-\lambda) d_{\text{bbox}}\end{equation*}
 where
\begin{equation*}d_{\text{Mahalanobis}}(x_i, \hat{x}_i, \Sigma_i) = \sqrt{(x_i - \hat{x}_i)^T \Sigma_i^{-1} (x_i - \hat{x}_i)}\end{equation*}
adds emphasis to reliable keypoints, \(\displaystyle \lambda\) is a weighting factor (0.8 worked best). 
The term $d_{\text{bbox}}$ measures the coordinate-wise difference between the 
projected keypoint bounding box and the segmentation mask bounding box:

\begin{align*}
d_{\text{bbox}} &= 
\big[\min(x_p),\, \min(y_p),\, \max(x_p),\, \max(y_p)\big] \\
&\quad - \big[\min(x_m),\, \min(y_m),\, \max(x_m),\, \max(y_m)\big]
\end{align*}

where $(x_p, y_p)$ are projected keypoint coordinates and $(x_m, y_m)$ are mask boundary coordinates. 
The residual calculates this error by projecting 3D points using current pose parameters, applying keypoint-specific covariance weighting, and comparing projected bounds with mask bounds.
This optimization is solved using \texttt{least\_squares} from SciPy, with specified tolerance parameters (\texttt{xtol=1e-8}, \texttt{ftol=1e-8}).

\subsection{Profile visibility analysis}
We add profile visibility information to the 3D bounding box label which is computed as described below. 

\subsubsection{Camera position and bounding box face normal}
We calculate the camera position $C_{pos} = -R^T t$ and then compute the outward face normals for the bounding box using the bounding box edges. 
For three adjacent vertices \( (v_0, v_1, v_3) \) of a face, normal $n$ is
\begin{equation*}
n = \text{normalize}((v_1 - v_0) \times (v_3 - v_0))
\end{equation*}
The \texttt{normalize} function, shown as
\begin{equation*}
\text{normalize}(\vec{v}) = \frac{\vec{v}}{\lVert \vec{v} \rVert}
\end{equation*}
where $\vec{v}$ is the input vector and $\lVert \vec{v} \rVert$ is its Euclidean norm, 
produces a unit vector with the same direction.
We ensure the normals are outward by
$n = -n \quad \text{if} \quad (n \cdot \text{direction}) < 0,$ where direction is calculated as the difference between the face center and the box centroid.

\subsubsection{Profile visibility and projected area}
We then compute the view vector $v$ from the camera to the face center using camera position $C_{pos}$ 
\begin{equation*}
v = \text{normalize}(C_{pos} - \text{face center})
\end{equation*}
and set a face visibility criterion through
$\text{visibility} = (n \cdot v) > 0
$.
For the visible faces, we calculated the projected area using Shoelace formula and then set a percentage using the total area of all visible faces. 
\begin{equation*}\text{Area} = \frac{1}{2} \left| \sum_{i=0}^{n-1} (x_i y_{i+1} - y_i x_{i+1}) \right|
\end{equation*}

\section{Evaluation}
We evaluated our 3D bounding box label generation pipeline on the Animal3D dataset and UAV-captured data. 
Qualitative results are shown in Figure \ref{fig:orientation_ablation}.  

\subsection{Orientation-aware 3D bounding box}
\begin{figure}[!htb]
  \centering
\includegraphics[width=0.9\linewidth]{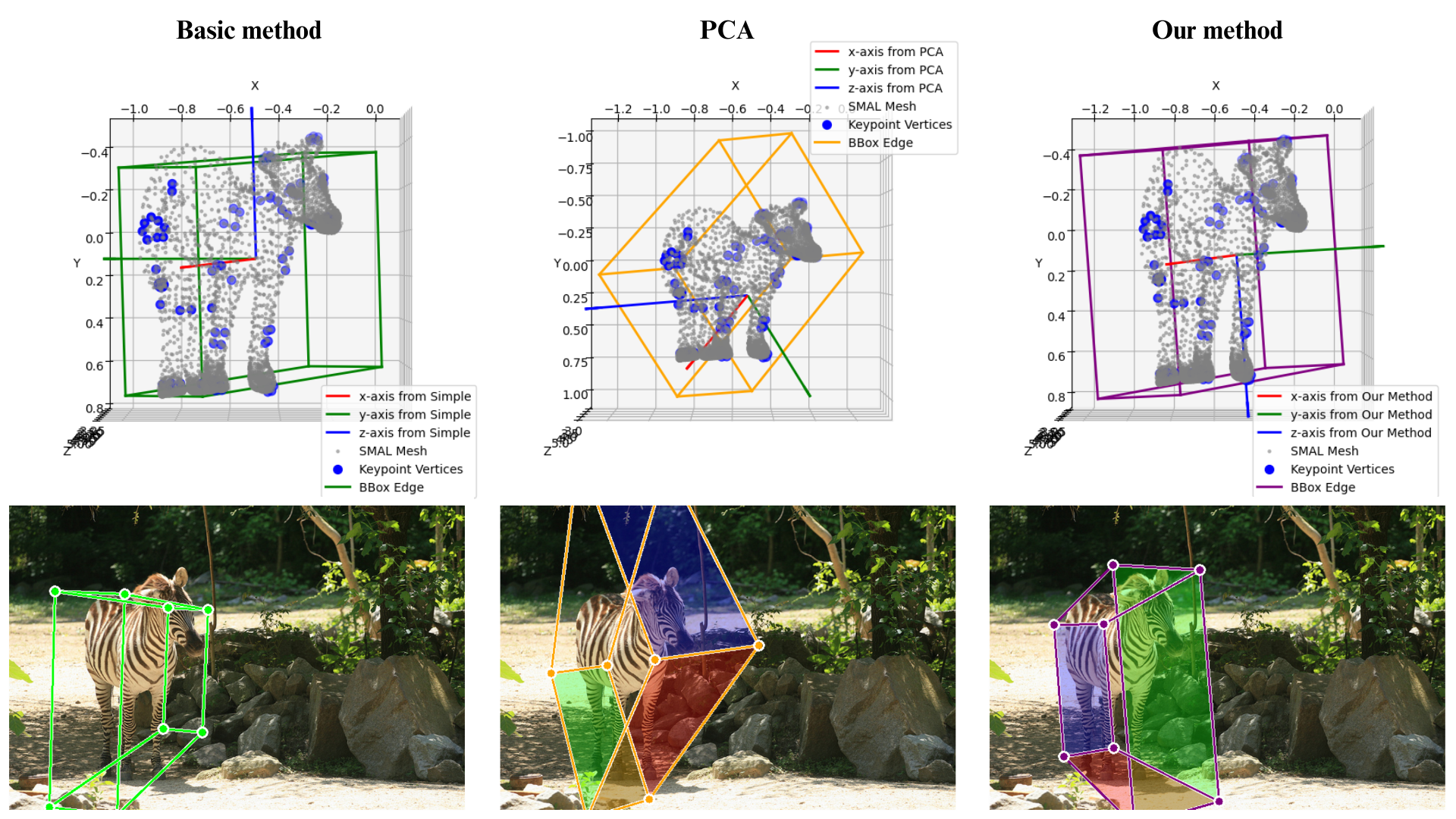}
   \caption{Visualization of oriented 3D bounding boxes on the SMAL mesh, along with their projections onto the corresponding images.}
\label{fig:orientation_qual_eval}
\end{figure}

Fitting a tight bounding box on the mesh with no orientation control gives tightest enclosed boxes in 3D but with arbitrary axis assignment and inconsistent reprojection. 
In most 3D labeling techniques for rigid objects, orientation is usually assigned with PCA, thus, we compare our method to PCA. 
PCA fails to generalize for axis calculation on animals, an example shown in Figure \ref{fig:orientation_qual_eval}. 
To test the stability of our landmark axis method, we introduce Gaussian noise 
$\mathcal{N}(0, \sigma^2)$ to the keypoint coordinates, where 
$\sigma \in \{0.5, 1.0, 2.0, 3.0, 4.0\}$ pixels represents different levels of 
perturbation. 
This range of noise levels was selected to simulate realistic 
keypoint localization errors that occur in practical detection systems. 
We then compare the rotation and alignment stability of the bounding box under these 
perturbations. 

Rotation variation $\theta_R$ is calculated in radians as
\begin{equation*}
    \theta_R = \arccos \left(\frac{\text{trace}(R_d) - 1}{2}\right)
\end{equation*}
where $R_d = R_1 R_2^{-1}$ is the rotation difference matrix between rotation matrices $R_1$ and $R_2$. 
For reporting purposes, we convert to degrees using $\theta_R^{\text{deg}} = \theta_R \cdot \frac{180}{\pi}$. 
This can be decomposed into anatomically meaningful components (anterior-posterior, left-right, dorsal-ventral), revealing that PCA exhibits particular instability in the anterior-posterior axis.
Our method shows a stability gain of over 99\% with the mean rotation variation in PCA being 120.68° and only 0.0046° with our method. 
This dramatic difference in stability arises from how each method responds to noise: PCA is highly sensitive to perturbations, as noise directly influences the computed principal components, especially for elongated structures like animal bodies, where small shifts in extremity positions can significantly alter the eigenvectors.
In contrast, our landmark-based approach relies on anatomically meaningful keypoints and their semantic relationships, which remain stable even when individual keypoint positions are perturbed within the expected range of detection errors ($\sigma \leq 4.0$ pixels).

We also calculate the alignment variation $\delta_a$, which measures the consistency 
between the ideal alignment vector and the actual alignment vector on noisy data. 
The ideal alignment vector $\mathbf{v}_{\text{ideal}}$ is derived from the original 
unperturbed keypoints by computing anatomical axes (anterior-posterior, left-right, 
and dorsal-ventral) based on landmark relationships. 
The alignment variation is given by:
\begin{equation*}
    \delta_a = 1 - \cos(\theta)
\end{equation*}
where $\theta$ is the angle between the ideal alignment vector $\mathbf{v}_\text{ideal}$ (e.g., front-back) and the alignment vector on noisy data $\mathbf{v}_\text{actual}$.
The mean alignment variation of the PCA method is approximately 0.5034 as compared to approx 0 for our method. 
A variation of 0.5034 indicates that the bounding boxes computed by the PCA method on noisy data have an alignment score that deviates significantly, approximately 50\% from the original, indicating lower stability.

\subsection{Camera pose refinement}
Our method significantly improves the reprojection results when compared to using simple 2D-3D keypoint correspondences which we will refer to as the basic method here. 
For Animal3D dataset, with a very strict degenerate threshold of 0.01\% i.e. bounding box area less than 1\% of the animal mask, the degenerate results are 13.81\% with the basic method. 
This goes down to 0\% with our method. 
The mean reprojection error goes down by 89.96\% from 75.28 pixels for the basic method to 7.56 pixels for our method.  

\subsection{Profile Visibility}
\begin{figure}[!htb]
  \centering
\includegraphics[width=0.95\linewidth]{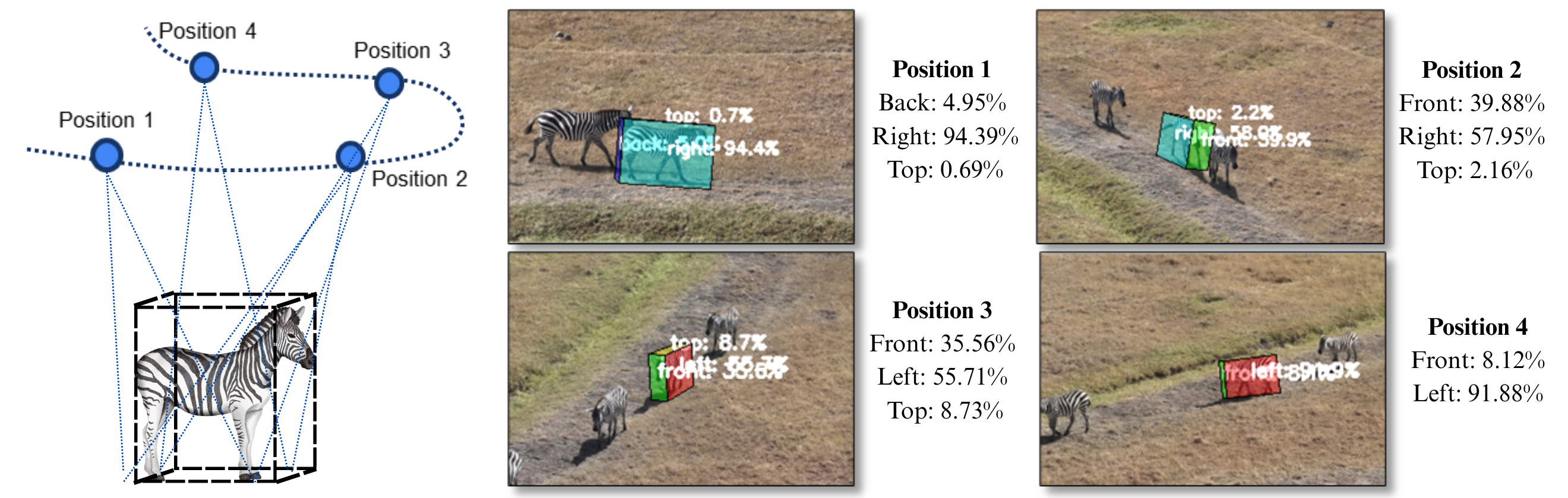}
   \caption{For each view on the zebra, we estimate the 3D bounding box along with the resulting visibility metric.}
   \label{fig:wd_visibility}
\end{figure}

We perform SMAL fitting on UAV-acquired data using the method in  \cite{shukla_towards_2024} and generate orientation-aware 3D bounding box labels to qualitatively assess the labeling method on aerial camera perspectives. 
UAV positions are illustratively shown in Figure \ref{fig:wd_visibility} with corresponding 3D bounding box and visibility results.

\section{Future Work}
Building on our 3D labels and visibility metric, we plan to release a dataset for monocular 3D animal detection along with benchmarks. 

\section{Acknowledgements}
The WildDrone project \url{https://wilddrone.eu/} has received funding from the European Union’s Horizon Europe research and innovation programme under the Marie Skłodowska-Curie grant agreement no. 101071224. We wish to thank the reviewers for their valuable input.
{
    \small
    \bibliographystyle{ieeenat_fullname}
    \bibliography{main}
}


\end{document}